\documentclass[final]{cvpr}

\usepackage{times}
\usepackage{epsfig}
\usepackage{graphicx}
\usepackage{amsmath}
\usepackage{amssymb}
\usepackage{color}
\usepackage{comment}
\usepackage{multirow}
\usepackage{subcaption}
\usepackage{wrapfig,lipsum,booktabs}
\usepackage{dsfont}

\DeclareMathOperator*{\argmin}{arg\,min}

\usepackage[pagebackref=true,breaklinks=true,colorlinks,bookmarks=false]{hyperref}

\def\cvprPaperID{4497} 
\def\confYear{CVPR 2021}

\begin{document}

\title{Uncertainty-Aware Physically-Guided Proxy Tasks for Unseen Domain \\ Face Anti-spoofing}


\author{%
 Junru Wu\textsuperscript{1}, Xiang Yu\textsuperscript{2}, Buyu Liu\textsuperscript{2}, Zhangyang Wang\textsuperscript{3}, Manmohan Chandraker\textsuperscript{2}\\
  {\textsuperscript{1}Texas A\&M University,
  \textsuperscript{2}NEC Laboratories America,
  \textsuperscript{3}University of Texas at Austin} \\
  \small{\texttt{sandboxmaster@tamu.edu,\{xiangyu,buyu,manu\}@nec-labs.com,atlaswang@utexas.edu}}
}




\maketitle

\begin{abstract}
Face anti-spoofing (FAS) seeks to discriminate genuine faces from fake ones arising from any type of spoofing attack. Due to the wide varieties of attacks, it is implausible to obtain training data that spans all attack types. We propose to leverage physical cues to attain better generalization on unseen domains. As a specific demonstration, we use physically guided proxy cues such as depth, reflection, and material to complement our main anti-spoofing (a.k.a liveness detection) task, with the intuition that genuine faces across domains have consistent face-like geometry, minimal reflection, and skin material. We introduce a novel uncertainty-aware attention scheme that independently learns to weigh the relative contributions of the main and proxy tasks, preventing the over-confident issue with traditional attention modules. Further, we propose attribute-assisted hard negative mining to disentangle liveness-irrelevant features with liveness features during learning. We evaluate extensively on public benchmarks with intra-dataset and inter-dataset protocols. Our method achieves the superior performance especially in unseen domain generalization for FAS.
\end{abstract}



\section{Introduction}
\label{sec:intro}

With growing prevalence of face recognition, it is increasingly subject to a wide variety of spoofing attacks. Thus, face anti-spoofing or liveness detection is emerging as an essential precursor, with the key need being robust to various attacks that are possibly unseen previously and drastically different in appearance from training data. This is an extremely challenging problem due to the fact that sophisticated spoofs might arise from similar camera and lighting setups as genuine inputs, leading to only subtle differences in appearance. On the other hand, types of attacks range from printed photos to facial masks, which makes it laborious to obtain exhaustive training data for anti-spoofing task.

\begin{figure}[t]
\centering
\includegraphics[scale=0.34]{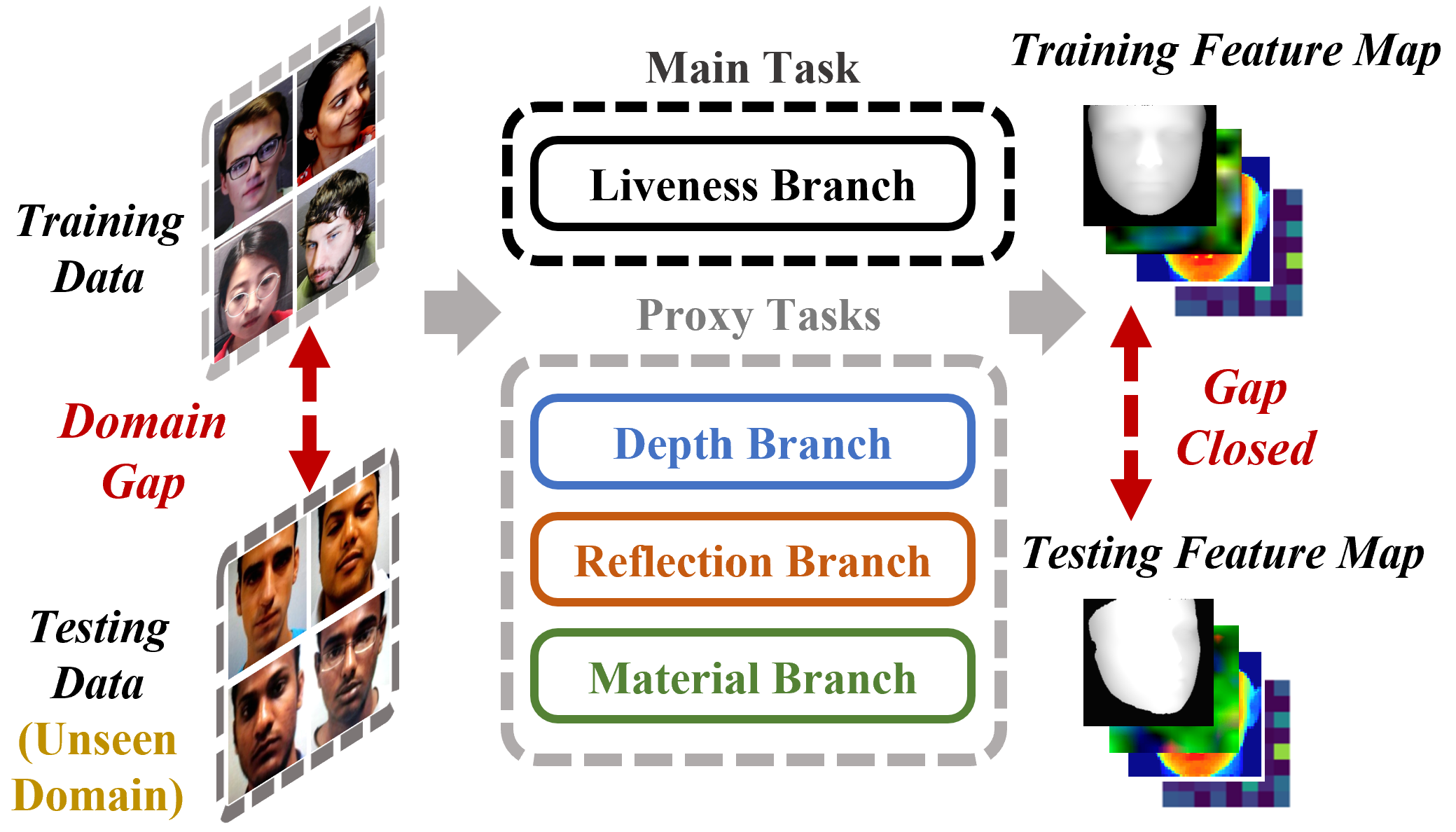}
\caption{Traditional methods only focus on real/spoofing binary classification, which results in sensitive prediction. By introducing the physical cues of depth, material and reflection as proxy tasks, our method largely close the domain gap from the training data to the unseen testing data and thus boost the performance reliably.}
\label{fig:teaser}
\vspace{-6mm}
\end{figure}

In this paper, we aim to address the face anti-spoofing problem on {\em unseen} domains or attack types, where neither definition of the attack types nor training data under supervised or unsupervised condition is available. To achieve this, we derive inspiration from physical cues that establish a commonality for genuine inputs and distinction from fake ones. We refer to the estimation of these cues as {\em proxy tasks}, performed in conjunction with the main task, a.k.a appearance-based liveness detection. While the proposed formulation is general and physical cues can be arbitrary, we focus on depth estimation, reflection detection and material classification as our proxy tasks. Intuitively, we expect genuine faces to constitute face-like geometry and present skin as the material, while several presentation attacks might violate at least one of those conditions. As a consequence, incorporating such proxy tasks enables to generalize the shared cues to unseen domains or attack types.

We bring the insights from single-image based face reconstruction using 3D morphable models (3DMM) \cite{Blanz999} for depth proxy, single-image based material recognition trained on large-scale datasets~\cite{bell15minc} for material proxy, and a single image reflection separation model~\cite{zhang2018single} to provide the pseudo labels for the reflection proxy. A shared encoder is trained across the main and proxy tasks to transfer the insights into our deep appearance-based liveness detection problem. In contrast to existing work~\cite{liu2018learning} that incorporates depth cue to regularize sensitive binary liveness task, our proxy tasks are more general in the sense of considering more physical cues such as material, to gear towards a physically meaningful way to handle unseen domains. Meanwhile, we organize the proxy tasks into a multi-channel learning framework to provide a more robust detection with an attention aggregation. Note that domain adaptation is not applicable in our setting, since we assume that even unlabeled training data is not available, which cannot define the target domain. 

Besides proxy tasks, we also leverage a {\em pretext task} in the form of face recognition, which is usually regularized by large scale labeled datasets and expected to provide high-level shared face analysis feature representation for liveness detection. We thereby provide recipes for pre-training on face recognition that allows better generalization to unseen domains for the liveness task, as well as multi-channel training with liveness and proxy tasks. We conduct extensive experiments on five publicly available benchmarks. In each case, we demonstrate not only state-of-the-art results, but also that judicious use of pretext and proxy tasks allows better generalization of liveness detection to unseen domains. Besides, the multi-task learning can result in channel conflict as the liveness feature is ideally invariant to identity information where the pretext task and our liveness task share the same network. To this end, we leverage the attribute information in the proxy data to conduct a triplet metric learning based mining, expecting to better disentangle the non-liveness information from the learned feature and thus boosts the liveness detection.

To better exploit multiple physically meaningful resources, we further holistically weigh the relative contributions of the main task and various proxy tasks with an uncertainty-aware attention module. Traditional attention modules are jointly optimized with all tasks and might cause the notorious over-confident issue due to training data bias. While our uncertainty-aware attention is designed to independently estimate the tasks' variance, which does not capture feature fitness to the task but rather focusing on its deviation to the estimated mean or termed uncertainty of the feature estimation. This property ensures less bias in uncertainty-aware attention module thus captures the property of input images better.

In summary, we propose the following contributions:
\vspace{-0.2cm}
\begin{itemize}
    \item{We propose three physical-cue guided proxy tasks including depth, material and reflection, which share the commonality across domains to enable the unseen domain anti-spoofing.}\vspace{-0.8em}
    \item{We leverage an uncertainty-aware attention module to effectively combine the main and proxy tasks and boost the performance.}\vspace{-0.8em} 
    \item{We propose a attribute-assisted mining scheme to make sure liveness-irrelevant features are properly disentangled and only liveness features are learned.}\vspace{-0.8em} 
    \item{We conduct an extensive evaluation with both intra-dataset and inter-dataset protocols including the latest attribute-rich CelebA-Spoof dataset, highlighting our framework's better performance in unseen domain generalization for FAS.}
\end{itemize}

\section{Related Work}

\begin{figure}[t]
\centering
  \begin{subfigure}[b]{0.23\textwidth}
    \centering
    \includegraphics[width=\textwidth]{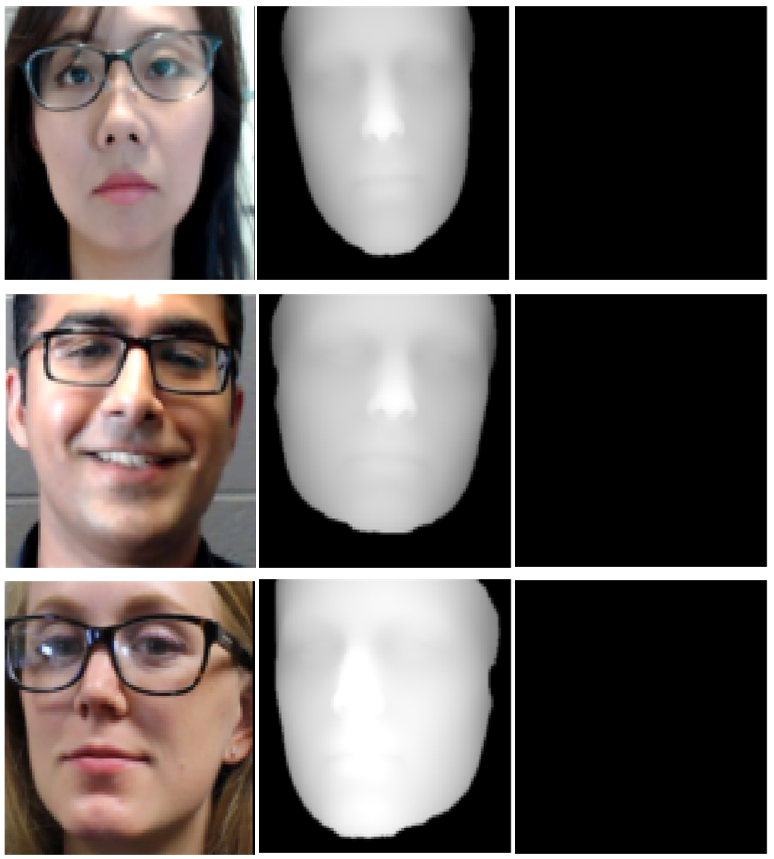}
    \caption{}
  \end{subfigure}
  \begin{subfigure}[b]{0.23\textwidth}
    \centering
    \includegraphics[width=\textwidth]{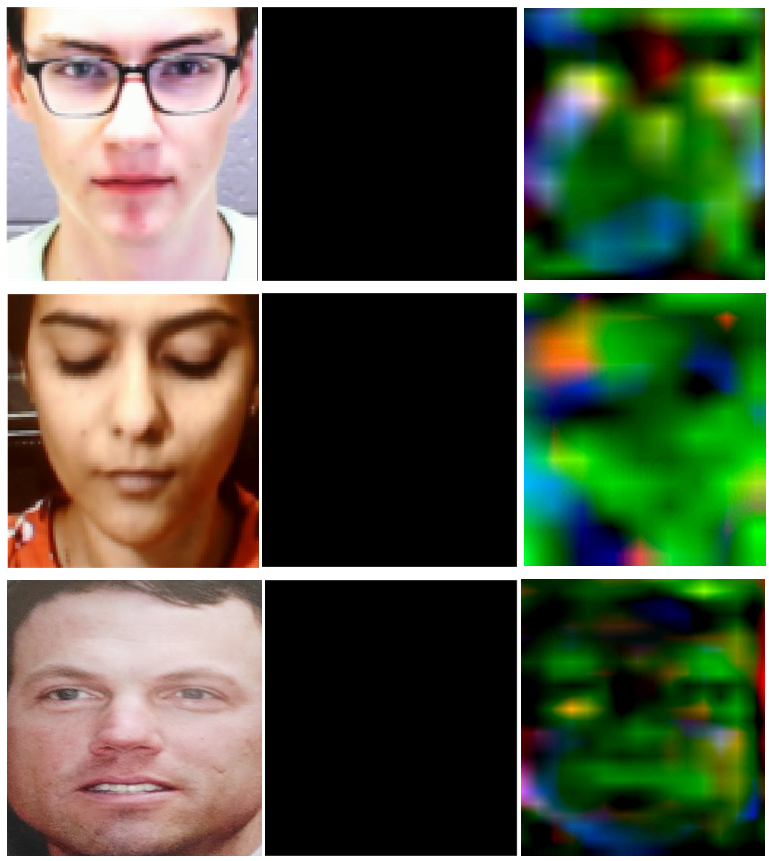}
    \caption{}
  \end{subfigure}
\vspace{-2mm}
\caption{(a) From Left to Right: examples of face image, depth map, reflection map for genuine faces (b) From Left to Right: examples of face image, depth map, reflection map for spoof faces.}
\vspace{-6mm}
\label{fig:depth}
\end{figure}

We categorize face anti-spoofing literature into physical cue based, feature learning based methods, and whether they address the unseen spoofing attacks.

\noindent\textbf{Physical Cue based Anti-spoofing:} Early research on anti-spoofing leverages the physical cues, e.g. head movement~\cite{kollreider2007} and eye blinking~\cite{eyeblink2007}, to indicate the genuineness. These methods can be simply spoofed by printing faces with eye region cut, or wearing a facial mask and moving head. By analyzing the lighting cooperated from different reflection, the remote Photoplethysmography (r-PPG)~\cite{hann_rppg_2013,bobbia_icpr_2016,rppgeccv16,nowara_fg_2017} is proposed to identify spoofing attacks with material information. However, this type of methods require the imaging quality to be high as the lighting measurement is less tolerated to noise. Combining with CNN, depth is proposed~\cite{liu2018learning,maddg2019} as an auxiliary task that enables less sensitive and more explainable training. \cite{kim2019basn} introduced reflection map as a supplement to depth for bipartite auxiliary supervision. This method is limited in generalizing to other spoofing resources since their depth, reflection and r-PPG are trained on a single dataset. Instead, our depth, reflection and material channels are guided by models trained on large scale datasets, i.e., a 3DMM based depth regression model~\cite{feng2018prn}, a single image reflection separation model~\cite{zhang2018single} and a material classification model~\cite{bell15minc}, which substantially improves the generalization.


\begin{table}[t]
\centering
\small
\scalebox{0.65}{
\begin{tabular}{lccccc}
 \hline
  Method & ~\cite{liu_dtl_2019} & ~\cite{patel_ccbr_2016} &~\cite{liu2018learning} & ~\cite{maddg2019} & Ours \\
 \hline
 Physical Cues & $\times$ & Blink & Depth, rPPG  & Depth & Depth, Reflection, Material\\
 \hline
 Temporal information & $\times$ & $\checkmark$ & $\checkmark$ & $\times$ & $\times$ \\
 \hline
 Multiple-domain data& $\checkmark$ & $\times$ & $\times$ & $\times$ & $\times$ \\
 \hline
 Cross-domain evaluation & $\checkmark$ & $\checkmark$ & $\checkmark$ & $\checkmark$ & $\checkmark$ \\
 \hline
 \end{tabular}}
 \caption{\small{Unseen domain anti-spoofing methods comparison with robust feature~\cite{patel_ccbr_2016}, anomaly detection~\cite{arashloo_access_2017}, binary or auxiliary supervision~\cite{liu2018learning}, deep tree learning (DTL)~\cite{liu_dtl_2019} and MADDG~\cite{maddg2019}. ``$\checkmark$'' means applying and ``$\times$'' means not applying.}}
 \label{tab:unseen_method}
 \vspace{-6mm}
\end{table}

\noindent\textbf{Learning based Anti-spoofing:} 
The handcrafted features, e.g., HoG~\cite{komulainen2013,yangicb2013}, SIFT~\cite{patel2016} and LBP~\cite{boulkenafet2015,pereira2012,jukka2011,wang2018exploiting,zhang2019dataset} are explored in early literature. Such binary classification achieves good performance but is restricted to some defined domains. Meanwhile, those methods do not consider environment variations, i.e., lighting, color tone or pose change. To this end, the HSV and YCbCr~\cite{boulkenafet2015}, Fourier transform~\cite{li_spie_2004} and image low-rank decomposition~\cite{tan_eccv_2010} methods are also explored. Some other works~\cite{agarwal2016,bao_iasp_2009,siddiqui2016,feng_jvcir_2016,xu_acpr_2016} utilize the temporal information assuming videos are available. Later, deep learning based features~\cite{yang_casia_2014,feng_jvcir_2016,li_ipta_2016,atoum_ijcb_2017,jourabloo_eccv_2018,liu2018learning,patel_ccbr_2016} are utilized~\cite{yang_casia_2014,feng_jvcir_2016,li_ipta_2016,atoum_ijcb_2017,jourabloo_eccv_2018,liu2018learning,patel_ccbr_2016} and achieve better performance. Notice that both \cite{atoum_ijcb_2017} and \cite{liu2018learning} leverage texture and depth, which seem to be close to our setting. However, instead of directly exploring texture, we formulate the texture into a more physically consistent cue, the material, and set up the material classification task to avoid rPPG calculation. Moreover, our method is single image based which does not require the temporal information, thus reducing the run-time and model complexity.

\noindent\textbf{Unseen Domain Anti-spoofing:} Methods that explore handcrafted features can deal with unseen spoofing attacks as these features are independent from attack types. However, due to the limitation of feature representation power, they cannot generalize well. A method comparison is listed in Table~\ref{tab:unseen_method}. Patel et al.~\cite{patel_ccbr_2016} propose to combine deep features with eye blinking cues for cross-dataset spoofing detection. There are works~\cite{arashloo_access_2017,xiong2018unknown} formulating the anti-spoofing task into an anormaly or outlier detection, which highly rely on the definition of genuine samples. To alleviate this, Liu et al.~\cite{liu_dtl_2019} propose a zero-shot learning solution, whereas the unseen attacks are assigned to the most similar attacks predefined in the database. These unseen attacks are wildly variant and leave the chance that the attacks are heavy outliers. Shao et al.~\cite{maddg2019} formulate a domain generalization approach to improve the generalization ability, which depends on the number and diversity of the seen domains, i.e., biased or long-tailed observed domains would degrade the performance. Different from \cite{liu_dtl_2019,maddg2019}, we propose the physical cue based proxy tasks that are less dependent on the data distribution, which generally could be more stable and consistent across seen and unseen attacks.

\noindent\textbf{Uncertainty Analysis:} Uncertainty provides an effective measurement for model/data reliability~\cite{lewis1994heterogeneous,sun2015active,kendall2017uncertainties,mukhoti2018evaluating}.
It has been widely applied in many vision tasks such as classification~\cite{joshi2009multi}, semantic segmentation~\cite{kendall2018multi} and face recognition~\cite{betta2011uncertainty}. Our method follows the setting of \cite{kendall2018multi} by leveraging multiple tasks, but in a complete different problem as face anti-spoofing rather than segmentation. We consider our proxy tasks are orthogonal to each other, whereas in \cite{kendall2018multi} those semantic tasks are strongly correlated. To the best of our knowledge, we are the first to leverage uncertainty in face anti-spoofing tasks.

\begin{figure*}[t]
\centering
\includegraphics[width=1.0\textwidth]{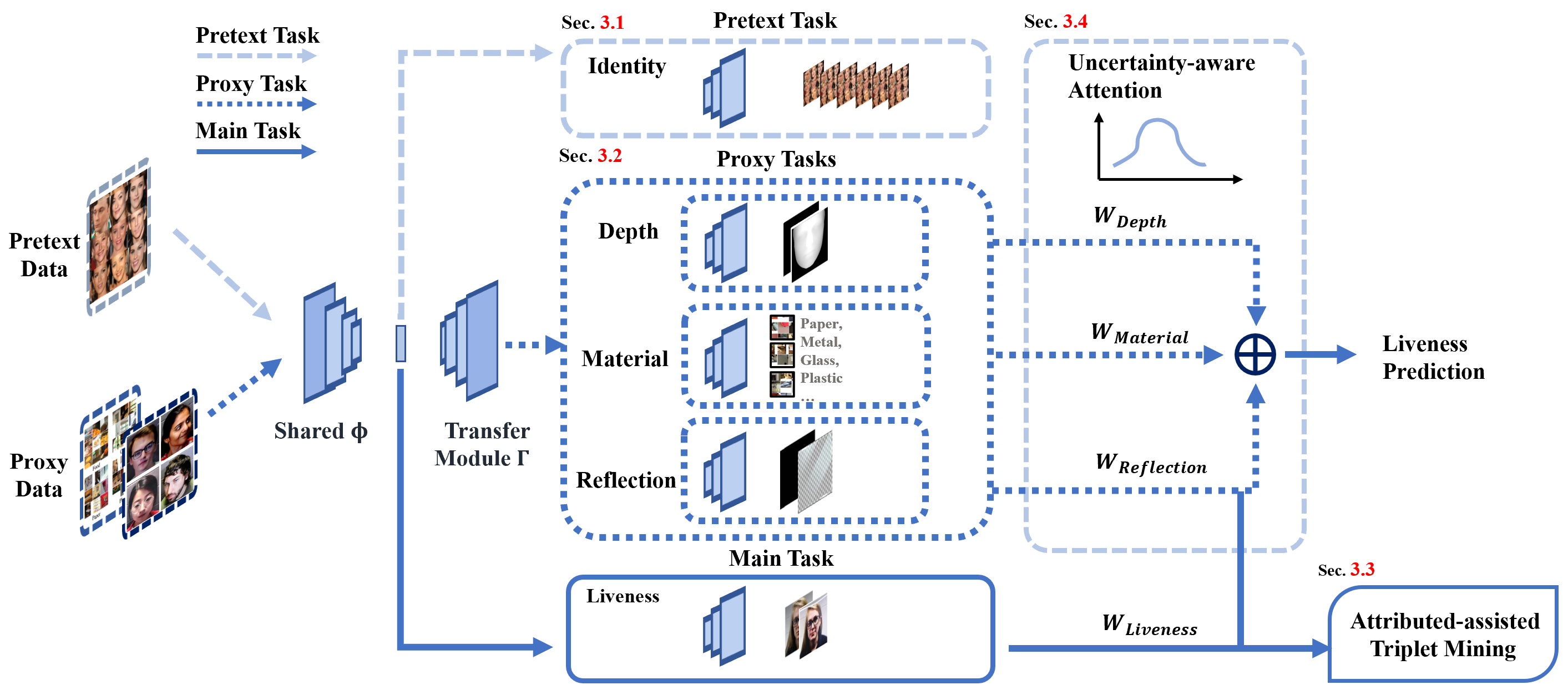}
\caption{The proposed framework consists of the pretext task ``face recognition'' (Sec.~\ref{sec:recognition}), the proxy tasks (Sec.~\ref{sec:proxy}) ``depth estimation'',``material prediction'', ``reflection detection'' and the main task ``liveness detection''. A novel triplet mining regularization (Sec.~\ref{sec:triplet}) is proposed to better disentangle the liveness feature and an uncertainty-aware attention module (Sec.~\ref{sec:attention}) aggregates the channel-wise results for boosted performance.}
\label{overview}
\vspace{-4mm}
\end{figure*}
\section{Proposed Approach}
In this section, we firstly introduce the shared feature extractor learning by incorporating the pretext task face recognition. Then, the physical cue based proxy tasks, i.e., depth estimation, reflection detection and material classification, are introduced as the spoofing attack detection anchors. Finally, an uncertainty-aware attention module is proposed to aggregate the proxy channels for optimal performance.

\subsection{Shared Feature Representation Learning} \label{sec:recognition}
As shown in Figure~\ref{overview}, our framework consists of multiple channels of pretext and proxy tasks. Separating each single task with independent CNNs results in network redundancy. Moreover, the separated CNNs cannot leverage the rich information from the other tasks, where hyper-column~\cite{hypercolumn2014} and deeply-supervised net~\cite{dsn2015} have shown a highly integrated framework for multiple tasks is beneficial. To this end, we propose to use a single feature extractor $\Phi$ to provide the shared feature for all the downstream tasks. 

The shared features should provide high-level task-specific yet general information for downstream tasks such that we neither drift away from original tasks nor learn only task-driven representations. Among the face analysis applications, face recognition is a promising pretext as it is usually trained with large-scale data including millions of identities, which guarantees the robustness as well as the discriminality. Other candidates such as facial attribute classification, expression recognition or spoofing detection are not general or robust, as each of the tasks conduct a 10-way or 2-way classification, which can be sensitive or easily overfitting~\cite{liu2018learning}. Thus, to initialize the feature extractor $\Phi$, we apply face recognition as our pretext task.

Denoting input image as $\mathbf{x}_r, \mathbf{x}_v$ and $\mathbf{x}_m$ for recognition, spoofing and material data respectively. After the shared feature extractor $\Phi$, the pretext task applies a filter $\Psi_r$ to refine the face identity feature. The loss is:
\begin{align}
    \mathcal{L}_{r}&=-\sum_i \mathds{1}(y_{i})\log\frac{\exp(\mathbf{w}_{i} \Psi_r\circ\Phi(\mathbf{x}_{r}))}{\sum_j \exp(\mathbf{w}_{j} \Psi_r\circ\Phi(\mathbf{x}_{r}))}
    \label{eq:recog}
\end{align}
where $y_i$ is the ground truth label for identity $i$. $j$ varies across the whole number of identities. $\mathbf{w}_i$ is the $i_{th}$ separation hyper-plane of the classifier. $\circ$ denotes the sequential network flow.

\subsection{Multi-channel Proxy Task Learning} \label{sec:proxy}
We introduce a transfer module $\Gamma$ to adapt the rich feature extracted by $\Phi$ into the spoofing detection related tasks. Directly utilizing the shared feature leads to sub-optimal prediction as it incorporates unrelated face recognition cues, which may serve as noise. Similar to the pretext task, we set up multiple channels for our proxy tasks, i.e., liveness detection $\Psi_{v}$, depth estimation $\Psi_{d}$, reflection detection $\Psi_{r}$ and material prediction $\Psi_{m}$.

\noindent\textbf{Liveness Detection Main Task:} The spoofing detection is a well-known binary classification task. The input is spoofing face $\mathbf{x}_{v}$. After shared extractor and feature transfer module, we set up the spoofing detection channel filter $\Psi_{v}$ to conduct the binary classification task, in which we adopt binary cross entropy loss as the objective:
\begin{align}
    \mathcal{L}_{v} &= -y_v\log(\mathbf{p}(\mathbf{z}))- (1-y_v)\log(1-\mathbf{p}(\mathbf{z}))
    \label{eq:cross_entropy}
\end{align}
\vspace{-6mm}
\begin{align}
    \mathbf{p}(\mathbf{z}) = \frac{\exp(\tilde{\mathbf{w}}_0\mathbf{z})}{\exp(\tilde{\mathbf{w}}_0\mathbf{z}) + \exp(\tilde{\mathbf{w}}_1\mathbf{z})}
    \label{eq:softmax}
\end{align}
where $\mathbf{y}_v$ is the ground truth of spoofing or genuine, $\mathbf{z}=\Psi_v\circ\Phi(\mathbf{x}_v)$ denotes the spoofing detection feature after the spoofing detection filter $\Psi_{v}$. $\tilde{\mathbf{w}}_0$ and $\tilde{\mathbf{w}}_1$ are the separation hyper-planes of the binary classifier. Likelihood of being spoofing sample $\mathbf{p}(\mathbf{z})$ is estimated via a softmax operation in Equation~\ref{eq:softmax}.

\noindent\textbf{Depth Proxy Task:} We believe the physical cues should share similar characteristics for genuine faces across different attack types or spoofing datasets. Thus the depth prediction should also be consistent. We aim to predict the per-pixel depth map given the input face image. We leverage an hourglass network structure to conduct this regression problem, which has been proved effective in key point detection~\cite{hourglass} and image segmentation~\cite{atrous_seg_2018}. To prepare the ground truth depth map $\mathbf{d}_{GT}$, we apply a 3D face shape reconstruction algorithm~\cite{feng2018prn} offline to estimate the dense point cloud for the face images. As for genuine face image, we utilize the predicted depth as its ground truth depth map, where background is set as $0$. For 2D spoofing face images, according to their attack types, i.e., display screen or paper, we know that the actual depth is from a flat plane of either screen or paper. Thus, we manually set the spoofing ground truth depth to be all $0$. The absolute depth is unnecessary since we only focus on the relative face geometry. We show some examples of generated depth map results in Figure \ref{fig:depth}.
During training, an $l_1$-based reconstruction loss is applied:
\begin{align}
    \mathcal{L}_{d} = \|\Psi_d\circ\Gamma\circ\Phi(\mathbf{x}_{v}) - \mathbf{d}_{GT}\|_1
    \label{eq:depth}
\end{align}
where $\Psi_d$ is the hourglass net depth estimation module, $\Gamma$ is the feature transfer module and $\mathbf{d}_{GT}$ is the ground truth depth. Notice that for depth estimation, we input the spoofing data $\mathbf{x}_v$ with the augmented ground truth depth map. We do not utilize extra depth data for this channel.

\noindent\textbf{Reflection Proxy Task:}
Reflection is another useful physical cue that indicates the genuine faces, as non-skin materials inevitably show abnormal reflection compared to skin. As a result, for spoofing face, we use a single image reflection separation model \cite{zhang2018single} to generate the reflection map, while for genuine face, we set the it to zero denoting no reflection is present. Visual examples of generated reflection maps are in Figure \ref{fig:depth}. During training, we push the predicted reflection map to be close to the pseudo ground truth under $l_1$-based constraint:
\begin{align}
    \mathcal{L}_{r} = \|\Psi_r\circ\Gamma\circ\Phi(\mathbf{x}_{v}) - \mathbf{R}_{GT}\|_1
    \label{eq:reflect}
\end{align}
where $\Psi_r$ is the hourglass net reflection estimation module
and $\mathbf{r}_{GT}$ is the ground truth reflection map.

\noindent\textbf{Material Proxy Task:} Though reflection in some way indicates the material information, we explicitly introduce material as another proxy task to leverage the correlation among the multiple tasks, expecting to benefit from the multi-task learning. The physical insight for material in liveness detection is that skin across different spoofing attacks or spoofing datasets should remain similar RGB information. We automatically obtain the material type for face spoofing data according to its attack type. For instance, we denote the material class of screen display and paper print as ``glass'' and ``paper'' respectively. In this way, we actually unify the material type towards the general material recognition~\cite{bell15minc}.

Notice that the number of material types in spoofing data can be limited, which may encounter the same sensitivity issue as the binary spoofing detection task. To this end, we introduce the general material recognition data to anchor the material feature space from being collapsed. Specifically, the general material recognition and our spoofing data material recognition share all the network structures except the last classifier layer. As in general material recognition, there are $23$ defined categories~\cite{bell15minc}, such as brick, metal, plastic, skin, glass, etc. We set up a $23$-way classifier $\mathbf{C}_g$ for the general material recognition and a $3$-way classifier $\mathbf{C}_v$ for our spoofing data material recognition. A multi-source scheme is proposed to train the modules of $\Phi$, $\Gamma$ and $\Psi_{m}$ jointly. Denoting the feature $\mathbf{f}=\Psi_m\circ\Gamma\circ\Phi(\mathbf{x})$, a combined multi-class softmax loss is applied to train $\mathbf{C}_g$ and $\mathbf{C}_v$:
\begin{align}
     \mathcal{L}_{m}&=-\sum_{i=1}^{23} \mathds{1}(l_i)\log\frac{\exp(\mathbf{\omega}_i \Psi_m\circ\Gamma\circ\Phi(\mathbf{x}_m))}{\sum_j \exp(\mathbf{\omega}_j \Psi_m\circ\Gamma\circ\Phi(\mathbf{x}_m))} -  \notag\\ 
     &\sum_{i=1}^{3} \mathds{1}(l_i)\log\frac{\exp(\tilde{\mathbf{\omega}}_i \Psi_m\circ\Gamma\circ\Phi(\mathbf{x}_v))}{\sum_j \exp(\tilde{\mathbf{\omega}}_j \Psi_m\circ\Gamma\circ\Phi(\mathbf{x}_v))}
    \label{eq:material_recog}
\end{align}
where $l_i$ is the material ground truth label, $\omega_i$ and $\omega_j$, $\tilde{\mathbf{\omega}}_i$ and $\tilde{\mathbf{\omega}}_j$ are the separation hyper-planes for $\mathbf{C}_g$ and $\mathbf{C}_v$ respectively. By alternatively feeding the material and spoofing data, we guarantee that $\mathbf{f}$ is generalized for not only the standard material recognition, but also the material recognition for face spoofing data.

\subsection{Attributed-assisted Triplet Mining} \label{sec:triplet}

To better disentangle the liveness feature apart from identity information and other facial attributes information, we leverage the metric learning to regularize the feature representation learning. Specifically, given the input $\mathbf{x}_v^{i}$, we would expect the following loss to be minimized such that the identity information can be decoupled from the liveness feature.
\begin{align}
\mathcal{L}_{tid} = \lfloor\|\Phi(x_v^{i,j}) &- \Phi(x_v^{i,k})\|^2 - \notag \\
&\|\Phi(x_v^{i,j}) - \Phi(x_v^{h})\|^2 + m_1\rfloor_{+}
\label{eq:tri_id}
\end{align}
$x_v^{i,j}$ means the $j^{th}$ liveness sample from identity $i$, while $x_v^{h}$ simply means the liveness sample from other identities as a negative sample.

Similarly for other face attributes introduced in CelebA~\cite{celeba,zhang2020celeba}, we believe the orthogonality can be preserved if those attribute information is disentangled from the liveness information. \begin{align}
\mathcal{L}_{ta} = \lfloor\|\Phi(x_v^{a_i,j}) &- \Phi(x_v^{a_i,k})\|^2 - \notag \\
&\|\Phi(x_v^{a_i,j}) - \Phi(x_v^{a_h})\|^2 + m_2\rfloor_{+}
\label{eq:ta_id}
\end{align}
$a_i$ indicates an attribute label and $a_h$ indicates a different attribute label for the negative sample. $m_1$ and $m_2$ here are the margin hyper-parameter set to squeeze the classification boundary for better feature learning.



\subsection{Uncertainty-aware Attention Modeling} \label{sec:attention}
As each of the channels looks into different aspects of the spoofing characteristics, we seek to combine those independent channels adaptively to boost the final spoofing detection performance. We introduce an uncertainty-driven attention module that is orthogonal to each of the main and proxy tasks, which thus effectively overcomes the over-confident issue of the traditional attention modules.

Given an input $\mathbf{x}_v$, the joint likelihood $p(y|\mathbf{x}_v)=p(\mathbf{z}|\mathbf{x}_v)p(\mathbf{d}|\mathbf{x}_v)p(\mathbf{r}|\mathbf{x}_v)p(\mathbf{f}|\mathbf{x}_v)$ according to the channel independence assumption, where $\mathbf{z}$ is from Equation~\ref{eq:softmax} as the main task feature, $\mathbf{d}=\Psi_d\circ\Gamma\circ\Phi(\mathbf{x}_{v})$ is from Equation~\ref{eq:depth} as the reflection feature,
$\mathbf{r}=\Psi_r\circ\Gamma\circ\Phi(\mathbf{x}_{v})$ is from Equation~\ref{eq:reflect} as the depth feature,
$\mathbf{f}=\Psi_m\circ\Gamma\circ\Phi(\mathbf{x})$ is from Equation~\ref{eq:material_recog} as the material feature. Maximizing the joint likelihood leads to maximizing the summation of each likelihood:
\begin{align}
\argmin -\log(p(y|\mathbf{x}_v)= -\sum_{\mathbf{u}=\mathbf{z},\mathbf{d},\mathbf{r},\mathbf{f}}\log(p(\mathbf{u}|\mathbf{x}_{v})
\end{align}

For each channel, we assume a Gaussian distribution $p(\mathbf{u}|\mathbf{x}_v) \sim \mathcal{N}(\mu_{\mathbf{u}}, \sigma_{\mathbf{u}})$ to capture the uncertainty, where $\mu_{\mathbf{u}}$ is the corresponding channel $\mathbf{u}$ classifier's separation hyper-plane vector or the mean depth map for the depth channel. Under the probabilistic setting, such $\mu_{\mathbf{u}}$ conforms to another Gaussian distribution $\mathcal{N}(\mu_{\mathbf{u}}, \sigma_{\mu_{\mathbf{u}}})$, where $\sigma_{\mu_{\mathbf{u}}}$ is estimated upon sampling from multiple rounds of training. $\sigma_{\mathbf{u}}$ is independently learned via a two FC-layer network structure in parallel to the feature $\mathbf{u}$. $\mu_{\mathbf{u}}$ is jointly learned with $\mathbf{u}$ during training and is fixed during the uncertainty training. The objective to learn $\sigma_{\mathbf{u}}$ is defined in the following:
\begin{align}
    \mathcal{L}_{\sigma_{\mathbf{u}}}=\sum_{\mathbf{u}=\mathbf{z},\mathbf{d},\mathbf{r},\mathbf{f}}\left(\frac{\|\mathbf{u} - \mu_{\mathbf{u}}\|^{2}}{2(\sigma_{\mathbf{u}}^{2}+\sigma_{\mu_{\mathbf{u}}}^{2})}+\frac{D}{2}\log(\sigma_{\mathbf{u}}^{2}+\sigma_{\mu_{\mathbf{u}}}^{2})\right)
    \label{eq:uncertainty}
\end{align}
where $D$ is the feature dimension. It is independently optimized after the network is converged. During inference, the network outputs $\mathbf{u}$ and $\sigma_{\mathbf{u}}$ simultaneously. Given $\mu_{\mathbf{u}}$ and $\sigma_{\mu_{\mathbf{u}}}$, we then obtain the uncertainty estimate for each channel with Equation~\ref{eq:uncertainty}.

To sum with, we propose a two-stage training procedure. The first stage consists of the training of liveness main task, proxy tasks and pretext task. And the loss is defined as following:
\begin{align}
    \mathcal{L}=\lambda_v\mathcal{L}_v + \lambda_d\mathcal{L}_d + \lambda_r\mathcal{L}_r + \lambda_m\mathcal{L}_m + \lambda_{t} (\mathcal{L}_{tid} + \mathcal{L}_{ta})
    \label{eq:overall_loss}
\end{align}
Then in the second stage, the uncertainty attention module is trained with Equation~\ref{eq:uncertainty}.


\section{Implementation Details}
In our implementation, we leverage a pre-trained face recognition engine and re-utilize the encoder as our shared feature extractor $\Phi$. Then, we keep the face recognition as our pretext task and equip the main and proxy tasks to form a multi-source multi-channel training. As illustrated in the methodology section, a two-stage training is conducted. For the first stage joint training of pretext, main and proxy tasks, we apply random cropping and horizontal flipping as data augmentation. 

We adopt Adam solver and the initial learning rate is set 0.0001. The momentum and weight decay are fixed as 0.9 and 0, respectively. Hyper-parameters in Equation~\ref{eq:overall_loss} is empirically searched via some hold-out validation as $\lambda_r=1,\lambda_d=1,\lambda_v=1,\lambda_m=0.1$, $\lambda_t=0.1$ for triplet need to add here together with $m_1$ and $m_2$ as in Equation~\ref{eq:tri_id} and~\ref{eq:ta_id} . respectively. For the second stage, when training the uncertainty-aware attention module, we re-use well-trained modules from the first stage, and only fine-tune the two-layer fully connected layers for each of the main and proxy tasks to estimate the variance. 

\begin{table*}[t]
\centering
\scriptsize
\scalebox{1.0}{
\begin{tabular}{ c | c  c  c | c  c  c | c  c  c }
     \hline
     Protocol & \multicolumn{3}{c|}{1} & \multicolumn{3}{c|}{2} & \multicolumn{3}{c}{3} \\
     \hline
     Methods & APCER & BPCER & ACER & APCER & BPCER & ACER & APCER & BPCER & ACER \\
     \hline
     $\text{SVM}_{\text{RBF}}$+LBP\cite{boulkenafet2017oulu}
     & 4.17 & 4.17 & 4.17
     & 5.29$\pm$4.39 & 5.29$\pm$4.39 & 5.29$\pm$4.39
     & 16.84$\pm$1.89 & 16.84$\pm$1.89 & 16.84$\pm$1.89 \\
     $\text{SVM}_{\text{RBF}}$+BSIF \cite{arashloo_access_2017}
     & 7.95 & 7.95 & 7.95
     & 7.34$\pm$3.30 & 7.34$\pm$3.30 & 7.34$\pm$3.30
     & 25.56$\pm$5.63 & 25.56$\pm$5.63 & 25.56$\pm$5.63 \\
     FAS-BAS\cite{liu2018learning}
     & 3.58 & 3.58 & 3.58
     & 0.57$\pm$0.69 & 0.57$\pm$0.69 & 0.57$\pm$0.69
     & 8.31$\pm$3.81 & 8.31$\pm$3.81 & 8.31$\pm$3.81 \\
     FAS-TD-SF\cite{wang2018exploiting}
     & 1.27 & 0.83 & 1.05
     & 0.33$\pm$0.27 & 0.29$\pm$0.39 & 0.31$\pm$0.28
     & 7.70$\pm$3.88 & 7.76$\pm$4.09 & 7.73$\pm$3.99 \\
     \hline
     Ours (L)
     & 0.66 & 0.66 & 0.66 &
     0.35$\pm$0.32 & 0.35$\pm$0.32 & 0.35$\pm$0.32
     & 7.97$\pm$5.03 & 7.97$\pm$5.03 & 7.97$\pm$5.03 \\
     Ours (L+D) & 0.47 & 0.47 & 0.47
     & 0.25$\pm$0.21 & 0.25$\pm$0.21 & 0.25$\pm$0.21
     &7.75$\pm$4.97 & 7.75$\pm$4.97 & 7.75$\pm$4.97\\
     Ours (L+M)
     & 0.57 & 0.57 & 0.57
     & 0.27$\pm$0.24 & 0.27$\pm$0.24 & 0.27$\pm$0.24
     & 7.80$\pm$4.95 & 7.80$\pm$4.95 & 7.80$\pm$4.95 \\
    Ours (L+D+R) 
    & 0.42 & 0.42 & 0.42
    & 0.27$\pm$0.22 & 0.27$\pm$0.22 & 0.27$\pm$0.22
    & 7.73$\pm$4.96 & 7.73$\pm$4.96 & 7.73$\pm$4.96 \\
    Ours (L+D+R+M) 
    & 0.44 & 0.44 & 0.44
    & 0.24$\pm$0.22 & 0.24$\pm$0.22 & 0.24$\pm$0.22
    & 7.52$\pm$4.91 & 7.52$\pm$4.91 & 7.52$\pm$4.91 \\
    \hline
    Ours (L+M+A) w/o Triplet
    & 0.56 & 0.56 & 0.56
    & 0.27$\pm$0.20 & 0.27$\pm$0.20 & 0.27$\pm$0.20
    & 7.773$\pm$9.91 & 7.773$\pm$9.91 & 7.773$\pm$9.91 \\
    \hline
    Ours (L+D+A)
    & 0.46 & 0.46 & 0.46
    & 0.25$\pm$0.22 & 0.25$\pm$0.22 & 0.25$\pm$0.22
    & 7.50$\pm$4.79 & 7.50$\pm$4.79 & 7.50$\pm$4.79 \\
    Ours (L+D+R+A) 
    & 0.40 & 0.40 & 0.40
    & 0.23$\pm$0.21 & 0.23$\pm$0.21 & 0.23$\pm$0.21
    & 7.43$\pm$4.82 & 7.43$\pm$4.82 & 7.43$\pm$4.82 \\
    Ours (L+D+R+M+A) 
    & \bf{0.36} & \bf{0.36} & \bf{0.36}
    & \bf{0.20$\pm$0.16} & \bf{0.20$\pm$0.16} & \bf{0.20$\pm$0.16}
    & \bf{7.32$\pm$4.80} & \bf{7.32$\pm$4.80} & \bf{7.32$\pm$4.80} \\
\end{tabular}}
\caption{Intra-dataset evaluation on SiW dataset. L: main spoofing/liveness detection. L+D: spoofing and depth channels. L+M: spoofing and material channels, L+D+M: spoofing, depth and material channels. L+D+R+M: spoofing, depth, reflection and material channels. L+D+A: spoofing and depth with attention. L+D+M+A: spoofing, depth and material with attention. L+D+R+M+A: our overall model with attention.}
\label{table:siw}
\vspace{-2mm}
\end{table*}

\begin{table*}[t]
\centering
\scalebox{0.8}{
\begin{tabular}{ c | c c c c | c c c c | c c c c | c c }
    \hline
    Dataset & \multicolumn{4}{c|}{CASIA} & \multicolumn{4}{c|}{Replay Attack} & \multicolumn{4}{c|}{MSU} & \multicolumn{2}{c}{All}\\
    \hline
    Attack Type & \multicolumn{1}{c}{V} & \multicolumn{1}{c}{C-P} & \multicolumn{1}{c}{W-P} & \multicolumn{1}{c|}{All}& \multicolumn{1}{c}{V} & \multicolumn{1}{c}{D-P} & \multicolumn{1}{c}{P-P} & \multicolumn{1}{c|}{All}& \multicolumn{1}{c}{P-P} & \multicolumn{1}{c}{H-V} & \multicolumn{1}{c}{M-V} & \multicolumn{1}{c|}{All} & Mean & Std \\
    \hline
    $\text{OC-SVM}_{\text{RBF}}$+IMQ\cite{arashloo_access_2017}
    & 63.26 & 59.43 & 66.81 & 63.34
    & 84.48 & 67.57 & 70.30 & 74.49
    & 53.94 & 84.75 & 76.56 & 72.61
    & 70.14 & 4.87\\
    $\text{OC-SVM}_{\text{RBF}}$+BSIF\cite{arashloo_access_2017}
     & 67.59 & 51.01 & \bf{96.33} & 72.76
     & 46.54 & 63.24 & 38.88 & 50.62
     & 62.06 & 80.56 & 64.06 & 69.25
     & 64.21 & 9.71\\
    $\text{SVM}_{\text{RBF}}$+LBP\cite{boulkenafet2017oulu}
     & 77.41 & \bf{87.14} & 69.48 & 77.61
     & 69.64 & 73.31 & 71.85 & 71.58
     & 55.39 & 96.02 & 94.88 & 83.36
     & 77.51 & 4.80\\
     NN+LBP\cite{xiong2018unknown}
     & 71.80 & 70.26 & 67.55 & 69.78
     & 36.93 & 75.43 & 69.45 & 59.75
     & 26.10 & 96.84 & 85.31 & 71.48
     & 69.75 & 8.30\\
     GMM+LBP\cite{xiong2018unknown}
     & 65.41 & 85.00 & 50.15 & 66.06
     & 60.78 & 61.46 & 55.32 & 59.57
     & 59.35 & 91.18 & 86.43 & 79.92
     & 68.51 & 8.48\\
     $\text{OC-SVM}_{\text{RBF}}$\cite{xiong2018unknown}
     & 64.94 & 85.75 & 55.15 & 67.95
     & 84.83 & 72.62 & 57.34 & 73.01
     & 60.90 & 68.41 & 75.51 & 68.60
     & 69.85 & 2.24 \\
     AE+LBP\cite{xiong2018unknown}
     & 77.72 & 80.30 & 52.92 & 69.56
     & 79.67 & 54.92 & 52.71 & 63.39
     & 55.67 & 87.94 & 92.18 & 79.67
     & 70.87 & 6.71 \\
     Auxiliary~\cite{liu2018learning}
     & - & - & - & 73.15 
     & - & - & - & 71.69 
     & - & - & - & 85.88 
     & 76.90 & 6.37 \\
     $^{*}$MADDG~\cite{maddg2019}
     & - & - & - & 84.51 
     & - & - & - & 84.99 
     & - & - & - & 88.06 
     & 85.85 & 1.57 \\
     \hline
     Ours (L)
     & 74.88 & 77.44 & 81.17 & 79.31     
     & 82.09 & 72.96 & 91.42 & 84.44
     & 66.25 & 96.60 & 95.34 & 85.81
     & 83.18 & 2.79\\
     Ours (L+D+A)
     & 80.02 & 81.79 & 87.80 & 87.61
     & 85.54 & 84.37 & 95.65 & 84.82
     & 67.82 & 97.52 & 96.16 & 88.59
     & 87.00 & 1.59\\
     Ours (L+D+R+A) 
     & 80.43 & 81.34 & 89.24 & 87.80
     & 86.15 & 84.56 & 95.63 & 85.23
     & 68.14 & 97.54 & 97.02 & 88.24
     & 87.09 & 1.32\\
     Ours (L+D+R+M+A) 
     & \bf{80.69} & 82.13 & 90.06 & \bf{87.92}
     & \bf{86.69} & \bf{84.92} & \bf{96.33} & \bf{85.27}
     & \bf{68.23} & \bf{97.70} & \bf{97.50} & \bf{89.23}
     & \bf{87.47} & 1.64\\
     \hline
  \end{tabular}}
  \caption{Inter-dataset evaluation on CASIA, Replay Attack and MSU, AUC(\%)  is reported. We follow the ``Leave one dataset \& attack-type out'' protocol in \cite{arashloo_access_2017}, where the attack types in testing set is unseen in the training set. We abbreviate V, C-P, W-P, D-P, P-P, H-V and M-V for Video, Cut Photo, Warped Photo, Digital Photo, Printed Photo, HR Video and Mobile video, respectively. *: retrained by their released codes.}
\label{table:inter-casia}
\vspace{-6mm}
\end{table*}
\vspace{-2mm}
\section{Experiments}

\subsection{Datasets}\label{subsec:datasets}



\noindent\textbf{CASIA}~\cite{zhang2012face}: A video based 2D spoofing attack database, consists of 600 videos from 50 people, 240 videos from 20 people for training and 360 videos from 30 people for testing. Each people contains 12 videos with video re-display and photo print attacks, of which 8 are normal resolution videos and 4 are high resolution videos. The photo attacks are further categorized into cut photo by cutting holes around eyes, noise, mouth, and warped photos by warping photos with different curvature.

\vspace{-0.5mm}
\noindent\textbf{Replay-Attack}~\cite{chingovska2012effectiveness}: A 2D face spoofing attack database consists of $1,300$ video clips of photo and video attack attempts from 50 clients, under different lighting conditions. To produce the attacks, high-resolution photos and videos from each client were taken under the same conditions as in their authentication step.

\vspace{-0.5mm}
\noindent\textbf{MSU-MFSD}~\cite{wen2015face}: it consists of 280 video clips of photo and video attack attempts from 35 clients. Mobile phones are used to capture both genuine faces and spoofing attacks. Printed photos are generated from high quality color printers for another attack type. It also provides replay video attacks with high resolution of $2048\times1536$ from iPad air screen.

\vspace{-0.5mm}
\noindent\textbf{Oulu-NPU}~\cite{boulkenafet2017oulu}: A large-scale 2D spoofing attack dataset, consists of 4950 genuine and attack videos from 55 people. They are recorded using the front cameras of different mobile devices with variant illuminations and backgrounds. The attack types are print and video replay. There are four protocols designed to consider generalization on cross attack types and capturing sensor types.

\vspace{-0.5mm}
\noindent\textbf{SiW}~\cite{liu2018learning}: Spoofing in the Wild dataset provides 4,478 genuine and spoofing videos from 165 subjects. For each subject, 8 genuine and up to 20 spoofing videos are captured. It systematically considers variations from subjects, camera sensors, spoofing attack types, lighting conditions, image resolution, and different sessions for capturing. There are three evaluation protocols emphasizing the generalization on face PAD, cross attack types, and unknown attack types.


\subsection{Evaluation Metrics}\label{subsec:metric}

The evaluation is focused on testing the generalization of cross attack types within one dataset, termed intra-dataset evaluation, and cross dataset spoofing detection, termed inter-dataset evaluation following \cite{arashloo_access_2017}.
To be consistent with most of the previous spoofing detection works, we apply the evaluation metrics as: Attack Presentation Classification Error Rate (APCER\cite{ISO_Biometrics}), Bona Fide Presentation Classification Error Rate (BPCER\cite{ISO_Biometrics}), ACER = 0.5(APCER+BPCER), Area Under Curve (AUC) ratio and Half-Total Error Rate (HTER). Further following~\cite{liu2018learning} in SiW protocol settings, we use Equal Error Rate (EER) as validation metric for all models to search the threshold to report performance.


\begin{figure*}[t]
\centering
\includegraphics[width=0.84\textwidth]{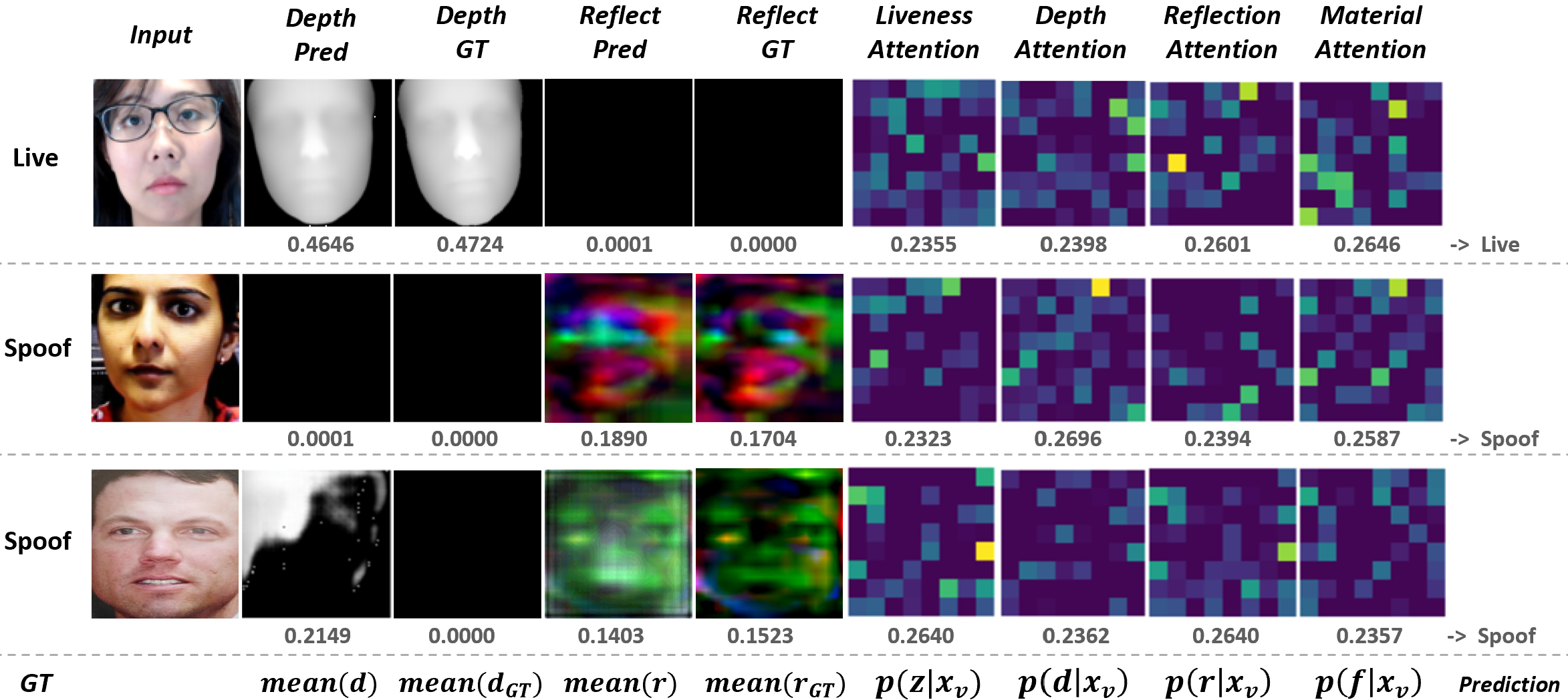}
\caption{Visual examples of our Uncertainty-aware Attention Modeling. We show example of Spoof and Live Image on SiW dataset. The first two rows shows examples is from SiW dataset while the bottom row is from CelebA-Spoof dataset~\cite{zhang2020celeba}.}
\vspace{-1em}
\label{fig:visual}
\end{figure*}

\subsection{Intra-Dataset Evaluation}\label{subsec:intra}
We evaluate on a recent large scale spoofing dataset SiW with carefully designed cross attack type testing protocols. We refer another intra-dataset evaluation on Replay-Attack to supplementary due to space limit.

\noindent\textbf{SiW Evaluation:}
There are 3 protocols in SiW. Protocol 1 focus on evaluating the performance of variations in face pose and expression. Protocol 2 focuses on the unseen medium of replay attack. It chooses 3 out of 4 display attacks, as training and leaving the remaining one as testing, which is iteratively conducted 4 times and averaged. Protocol 3 evaluates cross presentation attack detection, i.e., from print attack to replay attack and vice versa. Averaging over the two is reported.

In Table~\ref{table:siw}, our method consistently outperforms the other methods with significant margin, i.e., on Protocol 1, we achieve $\textbf{0.36}$ ACER while FAS-TD-SF~\cite{wang2018exploiting} is $1.05$. On Protocol 2, ours is $\textbf{0.20}$ while the best compared method is 0.31 from FAS-TD-SF. On Protocol 3, ours is $\textbf{7.32}$ while the best compared method is 7.73. Similar to Replay-Attack, we apply a gradually increasing module way to highlight effectiveness of the proposed modules. The ablation over our proposed modules suggests: (1) Depth, reflection and material are beneficial proxy tasks. (2) putting more proxy tasks together boosts the performance. (3) Our uncertainty-aware attention on top of the baselines can further achieve performance gain with significant margin. We also show an ablation contrasting w/ or w/o using attributed-assisted triplet mining, it shows that by adding triplet constraint, there is continuous margin gain over the other baselines.

\subsection{Inter-Dataset Evaluation}\label{subsec:inter}
The inter-dataset setting mimics the real setting for unseen attack across types and datasets. We consider two protocols. One follows the traditional rule~\cite{arashloo_access_2017}, a ``Leave one dataset \& attack-type out'' protocol, taking CASIA, Replay-Attack and MSU-MFSD as our datasets. Each of the three datasets contains three attack types. When evaluating one attack type of one dataset, we pick the other two datasets for training and excluding the testing attack type from training. The other less-strict setting is the ``Leave one dataset out`` protocol used in MADDG\cite{maddg2019}, the difference to the former is that in this protocol, training and testing sets would have overlapping attack types. 

\noindent\textbf{Leave one dataset \& attack-type out} In Table~\ref{table:inter-casia}, we evaluate each of the three attack types from three datasets. Both traditional feature learning based methods and most recent deep learning based methods~\cite{liu2018learning,maddg2019} are compared.
Overall we achieve consistently stronger results than the other methods. In CASIA, video attack is significantly better than other methods while cut photo and warped photo are among the top. In Replay-Attack, we achieve clear better performance. In MSU-MFSD, we observe 1\% to 6\% performance improvement over the compared methods.

\begin{table}[!t]
\centering
\scalebox{0.83}{
  \begin{tabular}{c | c | c | c | c | c}
     \hline
      Method & DD & \multicolumn{1}{c|}{OCI-M} & \multicolumn{1}{c|}{OMI-C} & \multicolumn{1}{c|}{OCM-I} & \multicolumn{1}{c}{ICM-O}\\
     \hline
     MS LBP\cite{maatta2011face} & \multirow{6}{*}{$\times$} & 78.50 & 44.98 & 51.64 & 49.31\\
     Binary CNN\cite{yang2014learn} & & 82.87 & 71.94 & 65.88 & 77.54\\
     IDA \cite{wen2015face} & & 66.67 & 39.05 & 78.25 & 44.59\\
     Color Texture \cite{boulkenafet2016face} & & 78.47 & 76.89 & 62.78 & 32.71\\
     LBPTOP \cite{de2014face} & & 70.80 & 61.05 & 49.54 & 44.09\\
     Auxiliary \cite{liu2018learning} & & 85.88 & 73.15 & 71.69 & 77.61\\
     \hline
     Ours (L) & \multirow{4}{*}{$\times$}
     & 85.81 & 79.31 & 84.44 & 76.26 \\
     Ours (L+D+A)
     & & 88.59 & 87.61 & 86.74 & 80.30 \\
     Ours (L+D+R+A)
     & & 90.73 & 89.03 & 88.42 & 83.58\\
     Ours (L+D+R+M+A)
     & & 91.32 & 89.28 & 91.83 & 85.48 \\
     \hline
     MADDG\cite{shao2019multi} & \multirow{3}{*}{$\checkmark$} & 88.06 & 84.51 & 84.99 & 80.02\\
     RFGML \cite{shao2020regularized} & & 93.98 & 88.16 & 90.48 & 91.16\\
     SSDG-M \cite{jia2020single} & & 90.47 & 85.45 & 94.61 & 81.83\\
     \hline
  \end{tabular}}
\caption{Inter-dataset evaluation on CASIA, Replay Attack, MSU and Oulu-NPU dataset. AUC (\%) is reported. We follow the ``Leave one dataset out'' protocol in \cite{maddg2019}, where training and testing sets share attack types. DD denotes disentangling source domains.}
\vspace{-4mm}
\label{table:maddg}
\end{table}
\noindent\textbf{Leave one dataset out} In Table~\ref{table:maddg}. Since our physically-guided proxy task does not require any domain priors, we compare methods w and w/o source domains disentanglement (DD). Our method surpass all methods without domain disentanglement including \cite{liu2018learning}, while still achieve comparable performance compare to methods \cite{shao2019multi}\cite{shao2020regularized}\cite{jia2020single} that utilize extra source domains information.
\subsection{Analysis of Uncertainty-aware Attention}
We visualize our Uncertainty-aware Attention Module in Figure~\ref{fig:visual}. Specifically, we visualize the last feature map of the last FC layer in the Attention Module across Liveness, Depth, Reflection and Spoof channel alongside with the corresponding Input Image, GT/Predicted Depth map and GT/Predicted Reflection map. In Figure~\ref{fig:visual}, we can see that, in most cases, those proxy tasks agree with each other like the examples from top two rows, while in some cases, those proxy tasks does not agree with each other. for example, the figure in the bottom row, depth channel made the wrong prediction and disagree with other channels, However, out uncertainty-aware attention module are able to correct his by give depth a lower confidence and voted for prediction from other channels, thus correcting the final decision.

\section{Conclusion}
In this work, we propose depth, reflection and material guided proxy tasks for unseen spoofing attacks. We propose a multi-source multi-channel training scheme for model optimization. Due to the consistency of depth, reflection and skin material across different spoofing scenario on genuine faces, by harnessing those physical proxy tasks, we expect the proposed method to deal with unseen spoofing attacks. Finally, an uncertainty-aware attention module is introduced to aggregate the multiple channels for boosted performance. Experiments across intra- and inter-dataset protocols show our method achieves consistently better performance and is effective for unseen spoofing detection.

{\small
\bibliographystyle{ieee_fullname}
\bibliography{egbib}
}

\end{document}